\documentclass{article}

\usepackage{arxiv}

\usepackage[utf8]{inputenc} 
\usepackage[T1]{fontenc}    
\usepackage{hyperref}       
\usepackage{url}            
\usepackage{booktabs}       
\usepackage{amsfonts}       
\usepackage{nicefrac}       
\usepackage{microtype}      
\usepackage{lipsum}		
\usepackage{graphicx}
\usepackage{natbib}
\usepackage{doi}

\usepackage{amsthm}
\usepackage{floatrow}
\usepackage{multirow}
\usepackage[misc]{ifsym}
\usepackage{float}
\floatstyle{plaintop}
\restylefloat{table}
\usepackage{siunitx}
\usepackage{tabularx}
\usepackage[english]{babel}

\usepackage{amsmath}
\newtheorem{definition}{Definition}[section]
\usepackage{amssymb}
\usepackage{subfigure}
\usepackage{booktabs}
\usepackage{multirow}
\usepackage{amsmath}
\usepackage{array,bm}
\newcolumntype{N}{>{\centering\arraybackslash}m{.5in}}
\newcolumntype{G}{>{\centering\arraybackslash}m{\dimexpr2in+6\tabcolsep}}
\usepackage{color}

\title{Curvature-based Pooling within Graph Neural Networks}

\author{ 
Cedric Sanders
	\And
\href{https://orcid.org/0000-0002-0515-7635}{\includegraphics[scale=0.06]{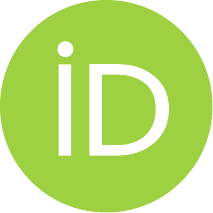}\hspace{1mm}Andreas Roth} 
	\And
	\href{https://orcid.org/0000-0002-9841-1101}{\includegraphics[scale=0.06]{orcid.pdf}\hspace{1mm}Thomas Liebig}
}

\author{ 
Cedric Sanders\textsuperscript{*} \\
	TU Dortmund University\\
	\texttt{cedric.sanders@tu-dortmund.de} \\
	\And
\href{https://orcid.org/0000-0002-0515-7635}{\includegraphics[scale=0.06]{orcid.pdf}\hspace{1mm}Andreas Roth\textsuperscript{*}(\Letter)} \\
	TU Dortmund University\\
	\texttt{andreas.roth@tu-dortmund.de} \\
	\And
	\href{https://orcid.org/0000-0002-9841-1101}{\includegraphics[scale=0.06]{orcid.pdf}\hspace{1mm}Thomas Liebig} \\
	TU Dortmund University,\\
Lamarr Institute for Machine Learning \\and Artificial Intelligence \\
	\texttt{thomas.liebig@tu-dortmund.de} \\
}

\renewcommand{\headeright}{}
\renewcommand{\undertitle}{}
\renewcommand{\shorttitle}{Curvature-based Pooling within Graph Neural Networks}

\hypersetup{
pdftitle={Curvature-based Pooling within Graph Neural Networks},
pdfsubject={machine learning, graph neural networks, pooling},
pdfauthor={Andreas Roth, Thomas Liebig},
pdfkeywords={machine learning, graph neural networks, pooling},
}

\begin{document}
\maketitle
\def\thefootnote{*}\footnotetext{Equal contribution\def\thefootnote{\arabic{footnote}}}
\begin{abstract}
Over-squashing and over-smoothing are two critical issues, that limit the capabilities of graph neural networks (GNNs). While over-smoothing eliminates the differences between nodes making them indistinguishable, over-squashing refers to the inability of GNNs to propagate information over long distances, as exponentially many node states are squashed into fixed-size representations. Both phenomena share similar causes, as both are largely induced by the graph topology. To mitigate these problems in graph classification tasks, we propose CurvPool, a novel pooling method. CurvPool exploits the notion of curvature of a graph to adaptively identify structures responsible for both over-smoothing and over-squashing. By clustering nodes based on the Balanced Forman curvature, CurvPool constructs a graph with a more suitable structure, allowing deeper models and the combination of distant information. We compare it to other state-of-the-art pooling approaches and establish its competitiveness in terms of classification accuracy, computational complexity, and flexibility. CurvPool outperforms several comparable methods across all considered tasks. The most consistent results are achieved by pooling densely connected clusters using the sum aggregation, as this allows additional information about the size of each pool.
\end{abstract}


\section{Introduction}
Graph neural networks (GNNs)~\cite{kipf16} combine the computational power of neural networks with the structure of graphs to exploit both the topology of graphs and the available graph signal. Their applications are manifold, as they are used to classify single nodes within a graph (node classification)~\cite{kipf16,roth2022forecasting}, classify entire graphs (graph classification)~\cite{zhang2018end}, and predict missing edges within the graph (link prediction)~\cite{pan2018adversarially}.
They are inhibited by several problems that impact the achieved results negatively. We propose a method to mitigate two of these problems for the graph classification task, namely over-smoothing~\cite{nt2019revisiting,oono2019graph,chen20} and over-squashing~\cite{alon20,topping21}.

Over-smoothing describes a phenomenon that results in node representations becoming overly similar when increasing the depth of the GNN. This leads to a loss of relevant information and leads to worse empirical results across many tasks~\cite{kipf16,li2018deeper,oono2019graph}. Various theoretical investigations confirmed that this problem is greatly enhanced by the underlying structure of the graph\cite{li2018deeper,oono2019graph,cai2020note}. Densely connected areas of the graph tend to over-smooth faster than sparsely connected areas~\cite{yan2022two}.

Similarly, over-squashing also leads to loss of information albeit in a different way. It describes the inability of GNNs to propagate information over long distances in a graph. A recent theoretical investigation traced this back to bottlenecks in the graph~\cite{alon20}, which describe edges connecting denser regions of the graph. With an increased number of layers, exponentially much information has to get passed through these edges, but the feature vectors are of limited constant size. Since bottlenecks are an inherent attribute of the underlying graph, this problem is also amplified by the graph topology~\cite{alon20,topping21}.

While various directions addressing over-smoothing have been proposed~\cite{topping21,roth2022transforming,yan2022two}, specifically for the graph classification task, pooling methods are a promising direction~\cite{ying18,luzhnica19}, which cluster sets of nodes in the graph into a single node. This can improve the data flow and change the underlying graph topology to one more suited for the respective task. The difficulty in applying pooling methods is the selection of these groups of nodes. Existing methods provide different criteria by which these nodes can be selected. Yet these are often prohibitively rigid and also not designed with over-smoothing and over-squashing in mind.

Our work addresses these crucial points. The curvature of a graph has been identified to be a meaningful metric for locating structures responsible for over-squashing and over-smoothing~\cite{topping21}. The curvature between two nodes describes the geodesic dispersion of edges starting at these nodes. Based on this metric, we design CurvPool, a novel pooling method that clusters nodes based on a flexible property of the graph topology. By design, the resulting graph has a suitable structure that alleviates the detrimental effects of over-squashing and over-smoothing. Our empirical results on several benchmark datasets for graph classification confirm the effectiveness of our approach. In addition, CurvPool is theoretically and practically efficient to execute.

\section{Preliminaries}
\paragraph{Notation}
We consider graphs of the form $G=(\mathcal{V},\mathcal{E})$ consisting of a set of $n=|\mathcal{V}|$ nodes $\mathcal{V} = \{v_1,\dots,v_n\}$ and edges indicating whether pairs of nodes are connected. For each node $v_i$, the set of neighboring nodes is denoted by $\mathcal{N}_i$ and its degree by $d_i=|\mathcal{N}_i|$. The graph signal $\mathbf{X}\in\mathbb{R}^{n\times d}$ consists of $d$ features at each node.
We consider the task of graph classification, which aims to find a suitable mapping $f_\theta(\mathbf{X},G) = c$ predicting class likelihoods $\mathbf{c}$ for the entire graph using some parameters $\theta$.

\subsection{Graph Neural Networks}
Graph neural networks operate on graph-structured data and are designed to extract meaningful node representations. These are structured as layer-wise functions to update the node representation
\begin{equation}
    \mathbf{h}_i^{k+1} = \psi(\mathbf{h}_i^{k},\phi(\{\mathbf{h}_j^{k} \mid j \in \mathcal{N}_i\}))
\end{equation}
in each layer $k$ using some neighbor aggregation function $\phi$ and some combination function $\psi$. The graph signal is used for the initial node representations $\mathbf{h}^0_i=\mathbf{x}^0_i$. Many options for realizing the update functions have been proposed~\cite{kipf16}.

However, most methods suffer from two phenomena known as over-smoothing~\cite{chen20} and over-squashing~\cite{alon20,topping21}. Over-smoothing refers to the case that node representations become too similar to carry meaningful information after a few iterations. Over-squashing occurs when exponentially much information is compressed into the representation of a few nodes, preventing information from flowing between distant nodes. This is induced by the structure of the graph as so-called bottlenecks cause over-squashing, which means that two parts of the graph are connected by relatively few edges.
\subsection{Pooling within GNNs}
A pooling operation reduces the spatial size of the data by aggregating nodes and their representations using some criterion.
It results in a new graph $G' = (\mathcal{V}^\prime,\mathcal{E}^\prime)$ and new node representations $\mathbf{H}^\prime$ with $|\mathcal{V}^\prime| \leq |\mathcal{V}|$.
Pooling methods offer various advantages which frequently include an increased memory efficiency and improved expressivity regarding the graph isomorphism problem~\cite{bianchi2023expressive}. 

Formally, each pool $p_i \subset \mathcal{V}$ contains a subset of nodes, and our goal is to find a suitable complete pooling
\begin{equation}
\label{eq_pooling} 
	\mathcal{P} = \{p_i \subset V \mid p_1 \cup \ldots \cup p_n = V\}
\end{equation}
so that every node of the graph is contained in at least one of the pools. This guarantees that no information that was contained in the initial graph is disregarded. The new set of nodes $\mathcal{V}^\prime$ is given by turning each pool $p_i$ into a new node $v_i^\prime$. 
The new set of edges $\mathcal{E}^\prime$ differs between pooling methods.
The main challenge towards successful pooling operations within GNNs is finding a suitable pooling criterion $\mathcal{P}$.

\subsection{The Curvature of a Graph}
\begin{figure}[tb]
    \centering
    \subfigure[]{\includegraphics[scale=0.15]{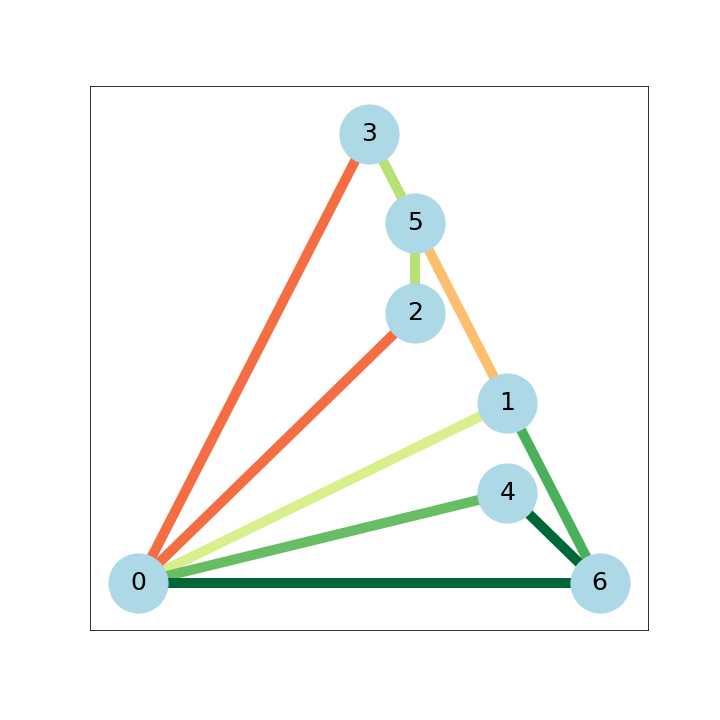}\label{fig_curvpool_a}}
    \subfigure[]{\includegraphics[scale=0.15]{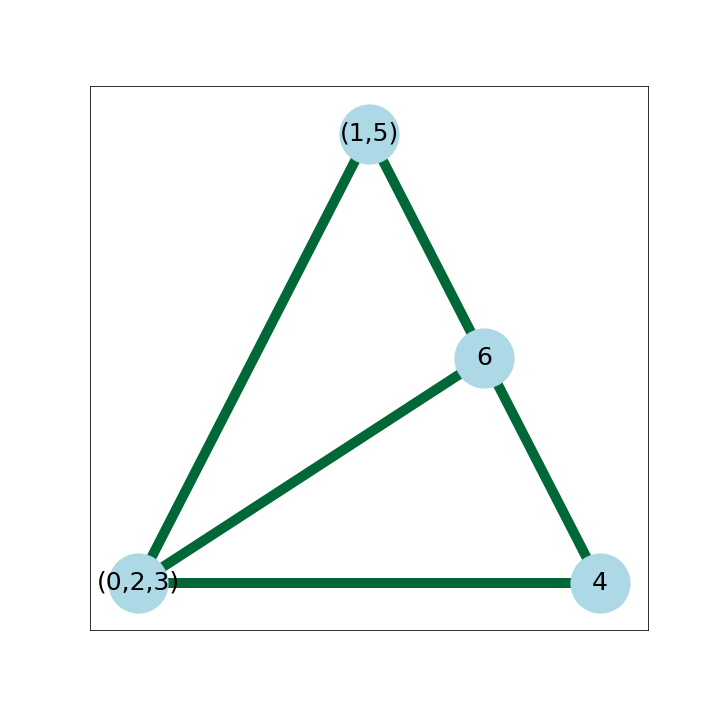}\label{fig_curvpool_b}}
    \subfigure[]{\includegraphics[scale=0.15]{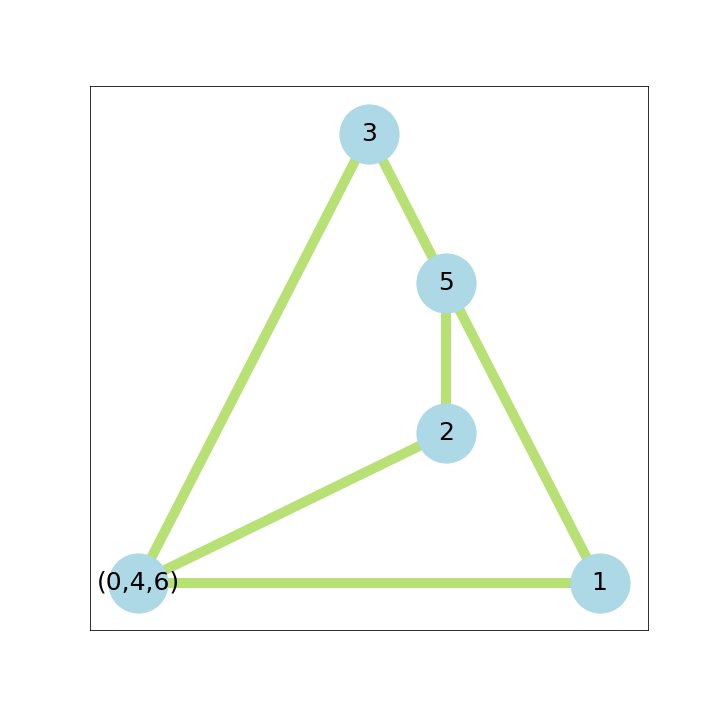}\label{fig_curvpool_c}}\\
    \subfigure[]{\includegraphics[scale=0.15]{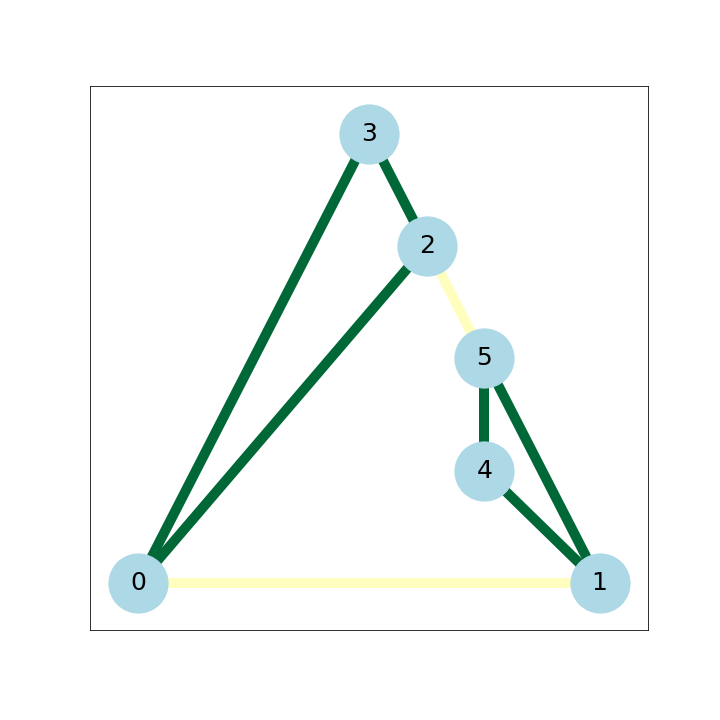}\label{fig_curvpool_d}}
    \subfigure[]{\includegraphics[scale=0.15]{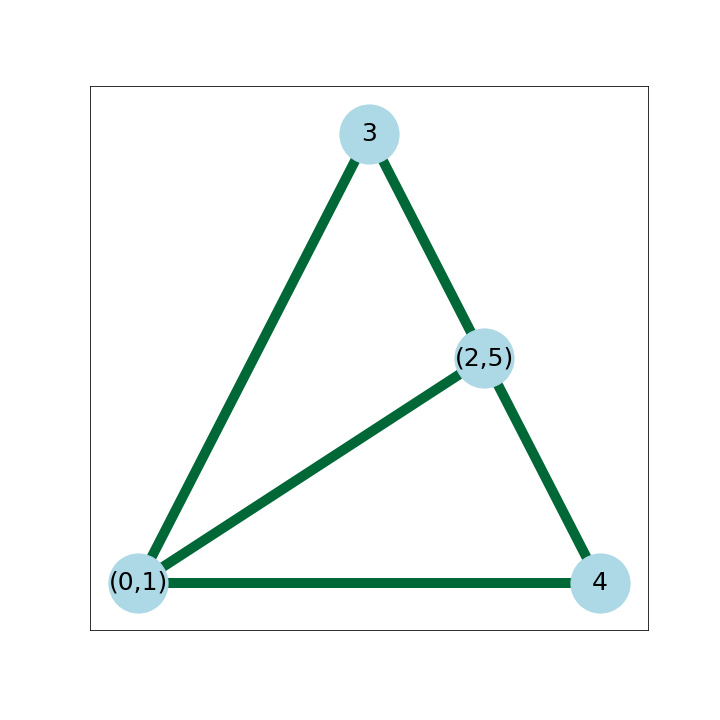}\label{fig_curvpool_e}}
    \subfigure[]{\includegraphics[scale=0.15]{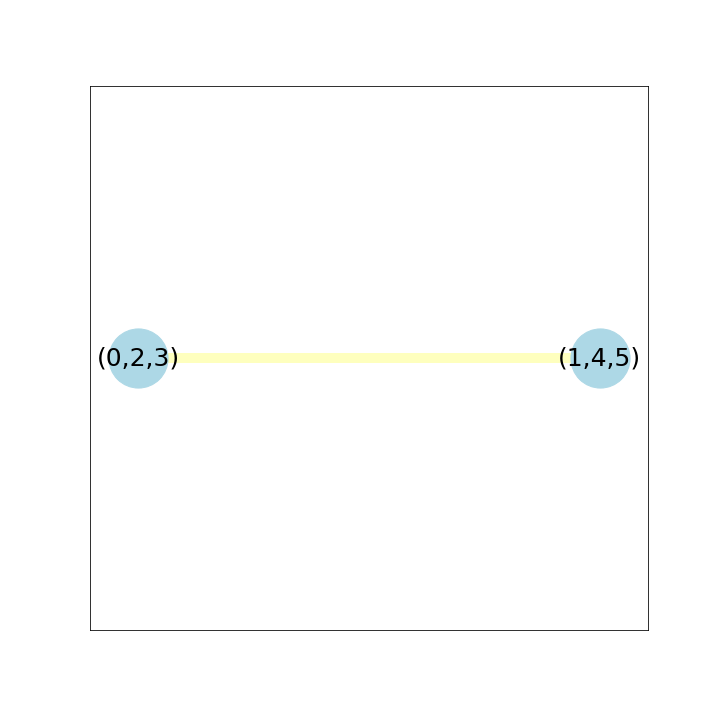}\label{fig_curvpool_f}}\\
    \caption[Example graphs \textit{CurvPool}]{Example graphs with edges colored according to their curvature from low (red) to high (green). Leftmost are the original graphs while the graphs in the middle represent one step of LowCurvPool and the rightmost graphs represent a step of HighCurvPool.}
        \label{fig_curvpool}
\end{figure}

Motivated by the Ricci curvature in Riemannian geometry, a recent investigation defined the curvature of a graph determines as the geodesic dispersion of edges starting at two adjacent nodes~\cite{topping21}. Two edges starting at adjacent nodes can meet at a third node, remain parallel, or increase the distance between the endpoints of the edges. Corresponding to these three cases and based on insights from previous edge-based curvatures~\cite{forman03,ollivier07,ollivier09}, they propose the Balanced Forman curvature:     

\begin{definition}
    (Balanced Forman curvature~\cite{topping21}.) For any edge $(i,j)$ in a simple, unweighted graph G, we let BFC(i,j) = 0 if $\min\{d_i,d_j\} = 1$ and otherwise
    \begin{equation}
        \textrm{BFC}(i,j) = \frac{2}{d_i} + \frac{2}{d_j} - 2 + 2\frac{|\triangle{(i,j)}|}{\max\{d_i,d_j\}} + \frac{|\triangle{(i,j)}|}{\min\{d_i,d_j\}} + \frac{\gamma_{max}^{-1}(i,j)}{\max\{d_i,d_j\}}(|\square^i| + |\square^j|)
    \end{equation}
    where $\triangle(i,j)$ are the 3-cycles containing edge $(i,j)$, $\square^i(i,j)$ are the neighbors of $i$ forming 4-cycles containing $(i,j)$ without containing a 3-cycle. $\gamma_{max}(i,j)$ is the maximal number of 4-cycles containing $(i,j)$ traversing a common node.
\end{definition}
We refer to Topping et al.~\cite{topping21} for a comprehensive definition.
This formulation satisfies the desired properties of the geodesic dispersion. The curvature is negative when $(i,j)$ are sparsely connected and positive curvature when redundant paths are available. We provide visualized examples in Figure~\ref{fig_curvpool}. 

The important relationship to over-squashing is that edges with a negative curvature are considered to be the bottlenecks of the graph~\cite{topping21}.

\section{Related Work}
Various methods for pooling within graph neural networks based on clusters of nodes have been proposed~\cite{yuan20,zhang18,ranjan20,lee19,wang20,gao19,diehl19,li20,nguyen18,zhang21,du21,lei22,zhao22}. 
Some are based on the topology of the graph, while others are based on the node representations themselves.

DiffPool~\cite{ying18} is one of the most frequently employed strategies for pooling based on representations. For each node, it predicts a soft assignment within a fixed number of clusters allowing the pooling to be optimized with gradient descent. Several other methods similarly learn a mapping from node representations to pools~\cite{noutahi2019towards,bianchi2020spectral,khasahmadi2020memory,liu2021hierarchical}. However, there are two main concerns with this family of strategies. First, the number of clusters is predefined and fixed for all graphs in the considered task. Second, the structure of the graph is only taken into account using node representations, which do not capture all structural properties, as given by their limitations regarding the Weisfeiler-Leman test~\cite{Morris_Ritzert_Fey_Hamilton_Lenssen_Rattan_Grohe_2019}.  

To address this, pooling strategies based on the graph topology were proposed. Fey et al.~\cite{fey2020hierarchical} predefine a fixed set of graph structures and pool only these into single nodes. CliquePool~\cite{luzhnica19} combines each clique in the graph. However, these methods rely on fixed structures in the graph and are unable to provide any pooling when the graph structure does not perfectly align. As an example, a graph could consist of densely connected communities, but these are only pooled when they constitute complete cliques.
For comparison throughout this work, we will use one of both categories, namely DiffPool and CliquePool.






\section{CurvPool}
We aim to construct an adaptive pooling method that can combine arbitrary structures in the graph without explicitly needing the knowledge of which structures we are interested in. In addition, the structure of the pooled graph should also be more resilient to over-smoothing and over-squashing, which then allows for a better flow of information. Using the curvature of the graph as the foundation for our pooling method allows us to achieve these properties.  

\subsection{Pooling based on the Curvature of a Graph}
\label{sec:method}
Since we base our new pooling approach on the Balanced Forman curvature $\textrm{BFC}(\cdot)$, we initially calculate the curvature for every edge $(i,j) \in E$. 
We then need to convert curvature values of edges to sets of nodes we want to pool together.
These values are used to decide if nodes $i$ and $j$ will be assigned to the same pool or not. There are different approaches for making this decision that lead to variations of CurvPool. In the general case, we use some criterion $f(\textrm{BFC}(i,j))$ on the curvature of each edge to decide whether two nodes are combined. For the initial candidates for pools, this results in a set
\begin{equation} 
	\mathcal{P}^\prime = \{\{i,j\} \mid (i,j)\in\mathcal{E}\land f(\textrm{BFC}(i,j))\} \cup \{\{i\} \in \mathcal{V}\}
\end{equation}
that we augment by each node as additional pool candidates to fulfill our requirements for a complete pooling. We describe our choices for the criterion $f$ in the next section.
The main challenge arises when combining subsets of $\mathcal{P}^\prime$. Nodes may be contained in multiple pools, as multiple edges of a node may satisfy our criterion for combination. This does not contradict our definition of a pooling but still should be considered since it significantly impacts the resulting graph. 
Intuitively, we want groups of nodes that are connected by edges of similar curvatures to be combined together. In this way, clusters of densely connected structures can be aggregated into a single node, and sparsely connected regions will be closer connected afterward.
The authors of CliquePool chose a different approach and removed duplicate nodes from every non-largest pool they were contained in~\cite{luzhnica19}. This approach doesn't really suit CurvPool since all resulting pools after the initial selection are of the same size. Instead, we merge all pools whose intersections are non-empty, resulting in the final pooling
\begin{multline} 
	\mathcal{P} = \{\bigcup_{p_i\in S} p_i \mid S\subseteq \mathcal{P}^\prime, \forall T \subset S: \left(\bigcup_{p_i\in T} p_i\right) \cap \left(\bigcup_{p_j\in S \setminus T} p_j\right) \neq \emptyset, \\ \forall p_k \in \mathcal{P}^\prime: \exists p_i\in S: p_i \cap p_k \neq \emptyset \Rightarrow p_k \in S\}.
\end{multline}
Each element in $\mathcal{P}$ is then mapped to a new node in the pooled graph. 
Based on the previous node representations $\mathbf{H} \in \mathbb{R}^{n\times g}$ of $\mathcal{V}$ and an aggregation function $\omega$, we construct new node representations 
\begin{equation}
 \mathbf{h}_{j}' = \omega(\{\mathbf{h}_i| i \in p_j\})
\end{equation}
for each pool $p_j\in\mathcal{P}$. Any aggregation scheme $\omega$ can be used to calculate the node features of the resulting pools. We consider the mean (AVG), the sum (SUM), and the maximum (MAX) operators. 

This still leaves us with one final question. How is the new set of edges calculated?
Since we strive to retain as much of the initial graph structure as possible, we simply remap the old edges from their respective nodes to the new pools they are contained in, resulting in
 \begin{equation}
 \mathcal{E}^\prime = \{ (m,n) \mid \exists\ p_i,p_j\in \mathcal{P}\colon p_i\neq p_j \land m\in p_i \land n\in p_j\}\, .
\end{equation}
This same method is used for all considered variations of CurvPool, leaving only the strategy for which curvatures to use for pooling open.

\subsection{Curvature-based Strategies for Pooling}
We now present our considered strategies for choosing pairs of nodes for our initial pools. Each of the three strategies has slightly differing motivations and carries its own set of advantages and disadvantages.

\subsubsection{HighCurvPool}
The fundamental idea of HighCurvPool is to aggregate nodes that are adjacent to edges with high curvature. This strategy combines all nodes that are connected by an edge with a curvature above a fixed threshold $t_{\mathrm{high}}$. Our initial set of pools
\begin{equation} 
	\mathcal{P}^\prime_{highCurv} = \{\{i,j\}\mid \forall i,j \in V: BFC(i,j) > t_{\mathrm{high}}\} \cup \{\{i\} \in \mathcal{V}\}
\end{equation}
considers exactly these, for which overlapping sets will be merged as described in Section~\ref{sec:method}.
Nodes are combined along the nodes in dense communities of the graph. As over-smoothing was shown to occur faster in dense communities~\cite{yan2022two}, these sets of nodes already contain similar representations, thereby being redundant when kept as separate nodes. HighCurvPool should alleviate this effect since the most strongly smoothed representations are aggregated, and the new graph contains more diverse neighboring states from each community. The effects of over-squashing should also reduce as the average path lengths become smaller and information from fewer nodes needs to be compressed for connecting edges.
While HighCurvPool typically leads to an increase in curvature in bottlenecks, these are not directly removed.

\subsubsection{LowCurvPool}
Analogous to HighCurvPool, LowCurvPool pools nodes that are connected by an edge with low curvature since these directly represent bottlenecks within the graph. Using a different threshold $t_{\mathrm{low}}$, this results in an initial pooling of the form
 \begin{equation} 
	\mathcal{P}_{lowCurv}^\prime = \{\{i,j\}\mid (i,j)\in\mathcal{E} \land \textrm{BFC}(i,j) < t_{\mathrm{low}}\} \cup \{\{i\} \in \mathcal{V}\}.
\end{equation}

LowCurvPool leads to the removal of exactly those edges that are marked as problematic through the curvature. The aggregation of the two adjacent nodes lets the two separated subgraphs move closer while guaranteeing that all paths through the graph are retained, and no new bottlenecks are created. As a result, the average curvature of the graph rises. Since we assume that the curvature is a good indicator of the over-squashing problem, information will be propagated better through the graph, and over-squashing gets reduced. However, over-smoothing may still be an issue as separate communities of nodes become more closely connected, leading to faster smoothing~\cite{yan2022two}. 



\subsubsection{MixedCurvPool}
Finally, MixedCurvPool combines the other approaches. It utilizes both thresholds $t_{\mathrm{high}}$ and $t_{\mathrm{low}}$. While one functions as an upper bound, the other works as the lower bound for our initial pooling
\begin{equation} 
	P_{mixedCurv}^\prime = \{\{i,j\}\mid \forall i,j \in V: BFC(i,j) < t_{low} \lor BFC(i,j) > t_{high}\}.
\end{equation}
Two nodes are combined along their edge either if the connecting edge represents a bottleneck or they are within the same densely connected community. The idea is that MixedCurvPooll combines the advantages of both approaches to be as effective as possible against over-squashing and over-smoothing. Selecting adequate hyperparameters becomes more important as this approach carries a great risk of simplifying the graph too much and thus losing all the information hidden inside the graph topology that we want to extract in the first place.

\subsection{Runtime Complexity}
\label{Complexity}
The runtime complexity of CurvPool is mainly given by the complexity of the Balanced Forman curvature. This complexity is $\mathcal{O}(|\mathcal{E}|d_{max}^2)$, with $d_{max}$ being the maximum node degree of the graph~\cite{topping21}. The calculation of the pools themselves only has a complexity of $\mathcal{O}(|\mathcal{E}|)$ while the complexity of merging overlapping pools is $\mathcal{O}(2|E|)$. All further operations don't differ between the pooling approaches and thus are not considered further for this comparison. As CurvPool only depends on the graph structure, this step is only executed once before optimization and reused in all settings.

CliquePools complexity is given through the calculation of the cliques via the Bron-Kerbosch algorithm~\cite{bron73} and is $\mathcal{O}(3^{n/3})$ in the worst case for a graph with n nodes \cite{tomita06}. This can frequently be reduced in a practical setting~\cite{eppstein13}.\\
Since DiffPool requires the calculation of a complete additional GCN its complexity is given by $\mathcal{O}(LN^2F + LNF^2)$ with $L$ being the amount of layers, $N$ being the amount of nodes and $F$ being the amount of features~\cite{blakely21}.

\section{Experiments}
To evaluate the effectiveness of our new pooling method, we compare its performance on different benchmark datasets on graph classification to well-established baselines. Our implementation is available online\footnote{\url{https://gitlab.com/Cedric_Sanders/masterarbeit}}.
We consider datasets that cover diverse graph structures and tasks from different domains. HIV~\cite{wu18} and Proteins~\cite{10.1093/bioinformatics/bti1007} are common benchmark datasets that are rooted within biology. They consist of structures of chemical substances that are used to classify the nature or specific properties of these substances. IMDB-BINARY~\cite{yanardag2015deep} is a common benchmark dataset that contains information about actors and movies. A graph indicates for a specific genre if actors, represented by nodes, have played together in the same movie, represented by edges. The classification task is to predict this genre. In addition, we extend our experiments to a custom dataset containing artificially generated graphs. These are generated using the approach for caveman graphs~\cite{watts99} resulting in multiple dense clique-like areas connected with a small number of edges that represent bottlenecks between these subgraphs. This dataset is referred to as Artificial.
This setup allows us to compare different CurvPool variations effectively as the distribution of high and low curvature areas within the graph is smooth. The used features are equivalent to the node degrees.

\subsection{Experimental setup}
The classification loss is calculated via the log-likelihood loss, while the used optimizer is the Adam-Optimizer using a learning rate of $0.001$ and a batch size of $32$. We employ a 10-fold cross-validation and choose the best setting based on validation accuracy. The best run across all hyperparameters is used to calculate the accuracy for held-out test data for each of the folds. All splits are consistent across models.

At each scale, our models utilize three convolutional layers, each followed by a ReLU activation and batch normalization~\cite{ioffe15}. 
The complete model consists of three of these blocks with the corresponding pooling layers in between. Since we focus on graph classification, a global mean pooling layer and two linear layers are used to calculate the final classification.
To keep them as comparable as possible, the pooling approaches only differ in the used pooling operation.

\subsection{Results}

Table~\ref{table:Ergebnisse_overall} presents the overall results for the experiments. It shows the best parameter constellation per dataset and method. 
The different variations of CurvPool outperform the established methods on almost all of the datasets, albeit usually by only a few percentage points. Only on the Proteins dataset DiffPool can keep up with CurvPool. Especially HighCurvPool outperforms all other considered methods consistently, with the second-best result typically also going to a variation of CurvPool.

\begin{table}[tb]
\setlength{\tabcolsep}{6pt}
\centering
\caption{Test accuracies for the best setting for each dataset and method according to the validation scores. The best score for each dataset is marked in bold, and the second-best score is underlined.}
\begin{tabular}{lcccccccc}
\toprule
\multirow{2}{*}{Dataset} & \multirow{2}{*}{GCN} & \multirow{2}{*}{DiffPool} & \multirow{2}{*}{CliquePool} & \multicolumn{3}{c}{CurvPool} \\
\cmidrule(lr){5-7}
&&& & Mixed & Low & High\\
\midrule
HIV&	$75.83$ &76.66	& 78.33& 76.31 & \underline{78.61} & \textbf{80.06}\\
Proteins&		65.99&\textbf{77.81}	&74.54		& 75.27	& 75.09	& \textbf{77.81}\\
IMDB-BINARY	&63.80	& 69.60	&68.39	&\textbf{70.80} & 69.40	& \textbf{70.80}\\
Artificial& 73.20	&73.80	&73.40	& 73.30 & \textbf{74.10}	&\underline{74.00}\\
\bottomrule
\end{tabular}
\label{table:Ergebnisse_overall}
\end{table}

\subsection{Ablation Study}
\begin{figure}[tb]
    \centering
    \subfigure[\textit{HIV}]
          {\includegraphics[scale=0.075]{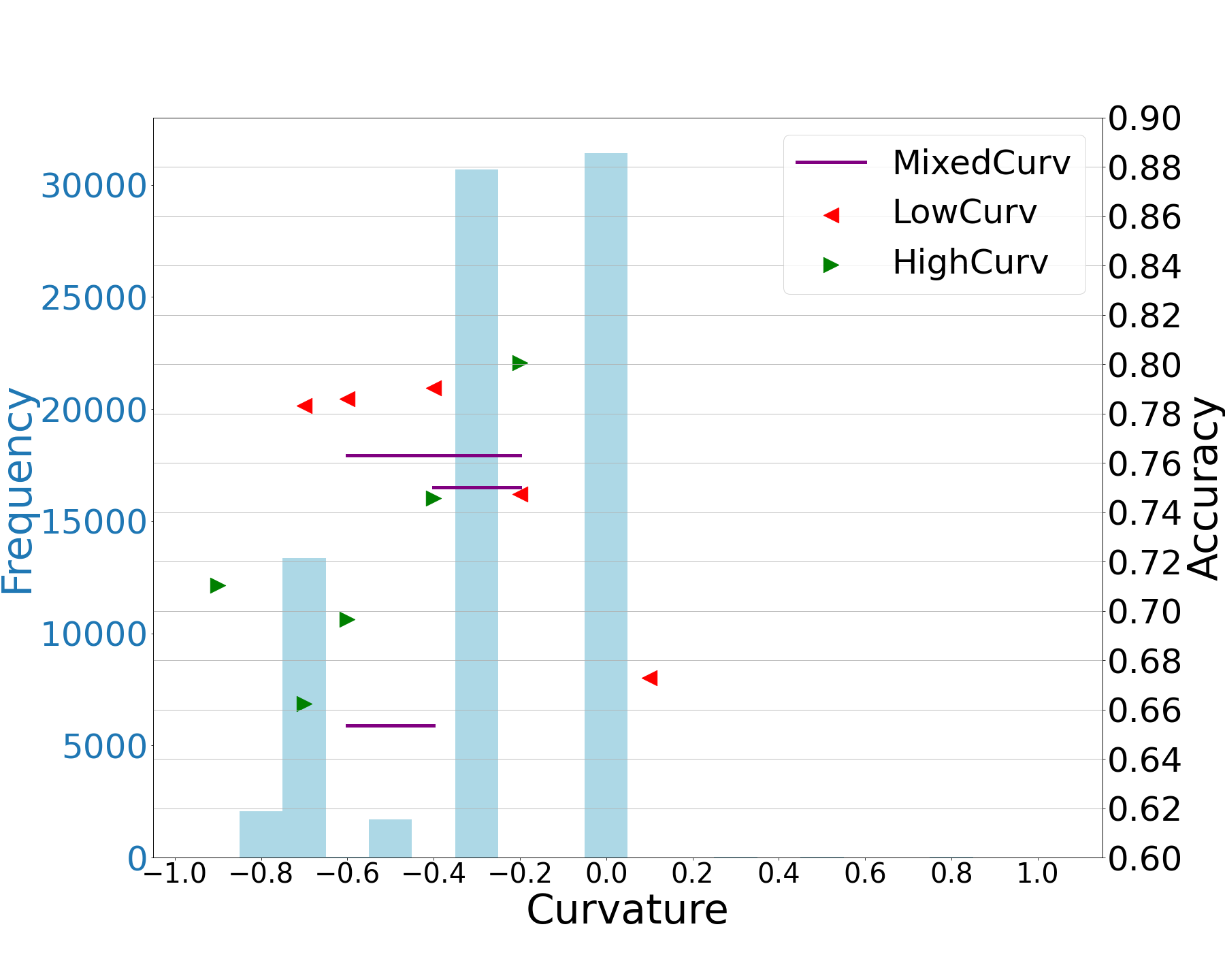}\label{fig_hiv_var}
    }
    \subfigure[\textit{Proteins}]
         {\includegraphics[scale=0.075]{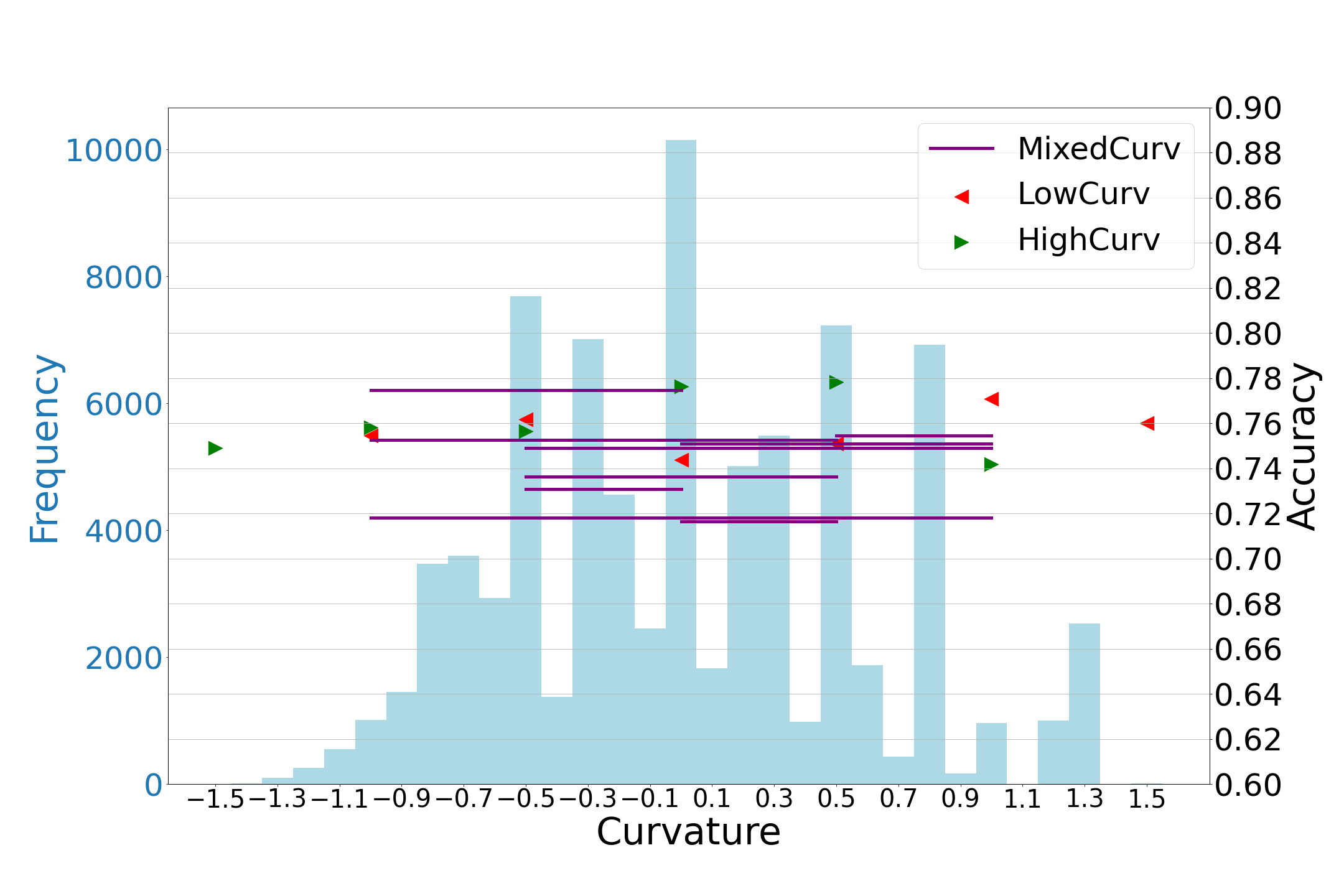}\label{fig_proteins_var}
    }
    \subfigure[\textit{IMDB-BINARY}]
         {\includegraphics[scale=0.075]{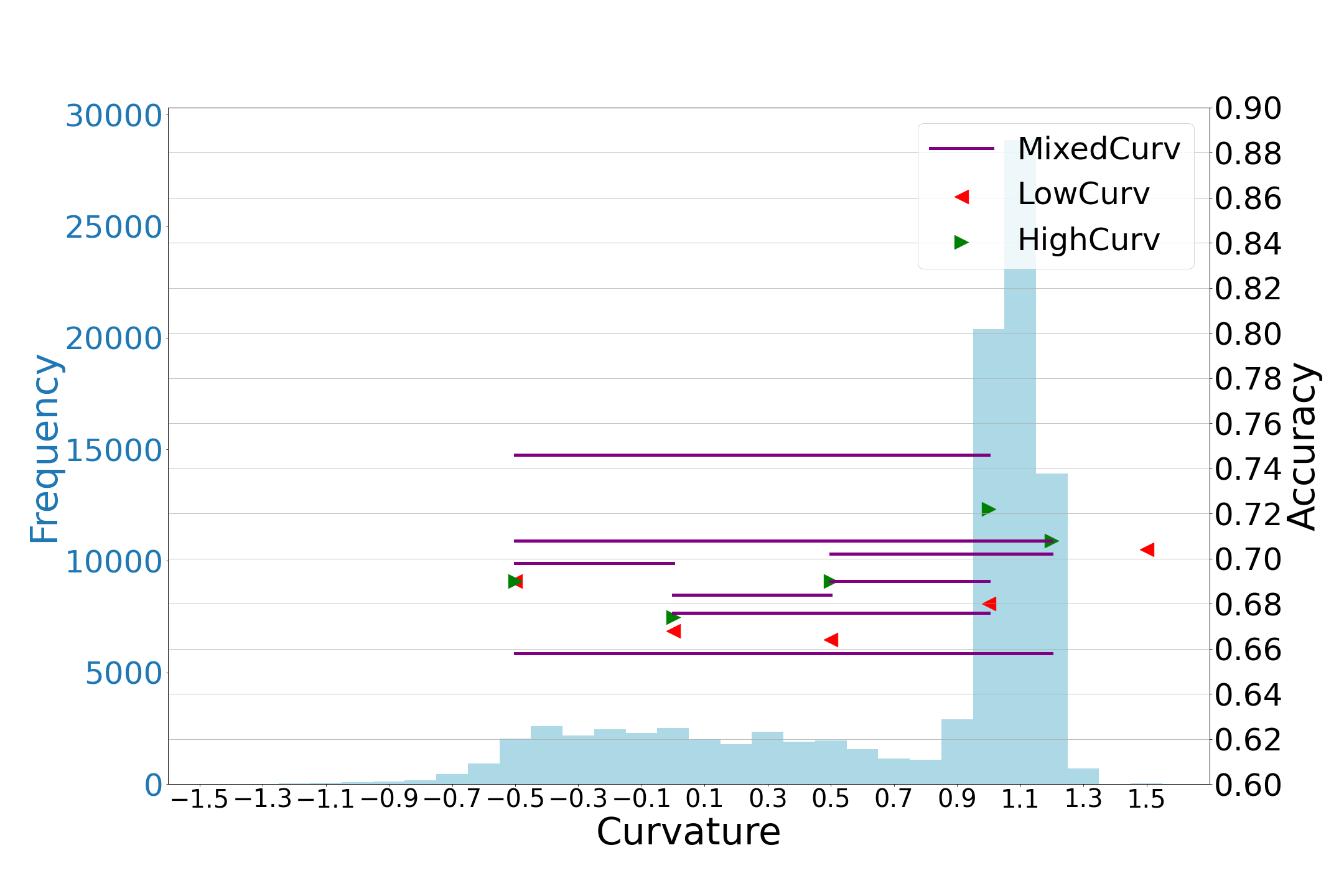}\label{fig_imdb_var}
    }
    \subfigure[\textit{Artificial}]
         {\includegraphics[scale=0.075]{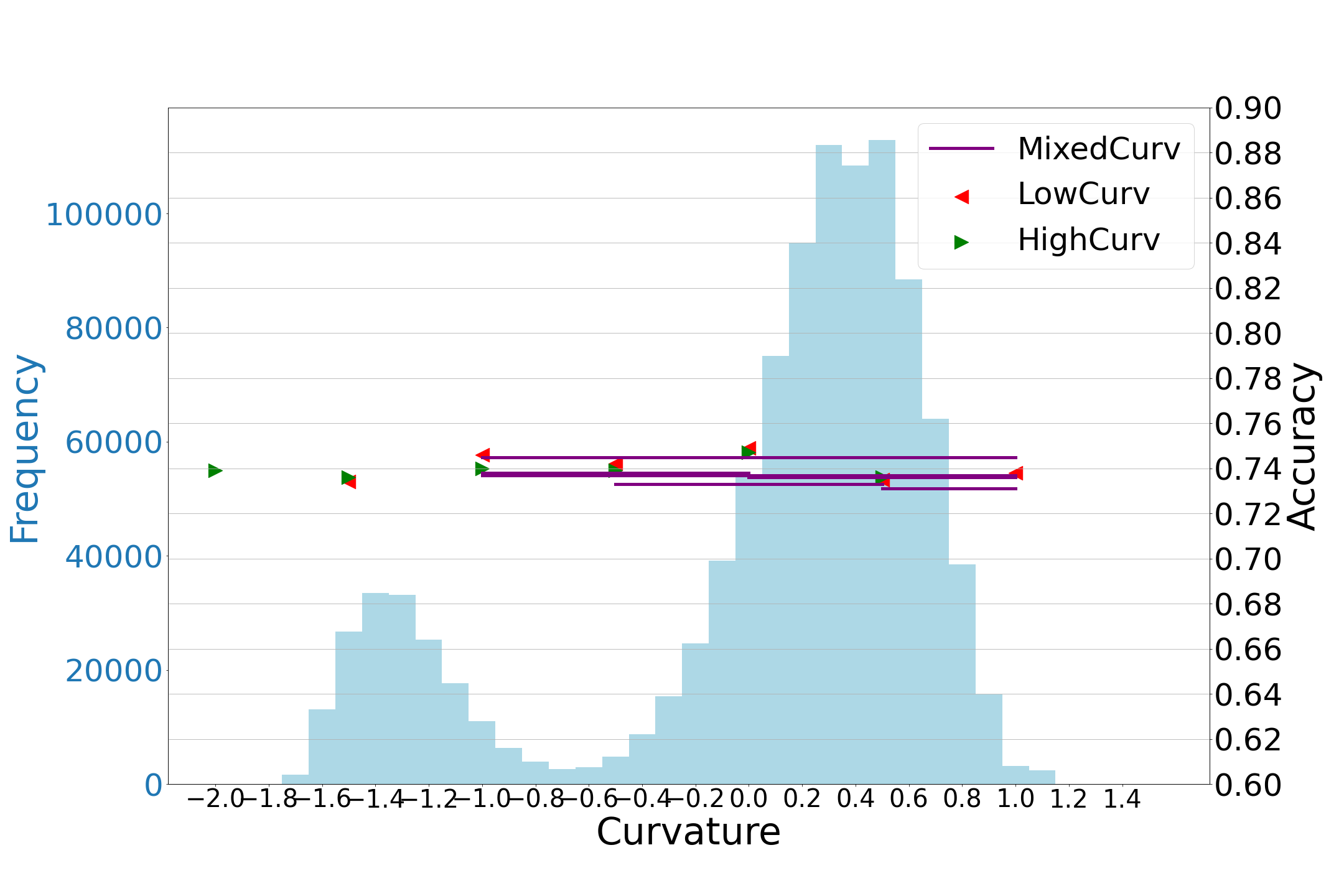}\label{fig_artificial_var}
    }
    \caption{Accuracy on the datasets for different thresholds. The histogram represents the distribution of curvature values in the dataset. MixedCurv pools all data points not within the range given via the purple line. LowCurv pools all data points to the left of the red triangle. HighCurv pools all data points to the right of the green triangle. }
        \label{fig_data_hist}
\end{figure}

The next step is to take a closer look at how some of the parameters impact the results of CurvPool. First up are the thresholds. Figure~\ref{fig_data_hist} represents the accuracy scores for the different sets of thresholds, including the different CurvPool variations. To better understand the impact of the thresholds the given histogram represents the distribution of curvature values in the corresponding dataset. This kind of visualization also allows for a good comparison of LowCurvPool, HighCurvPool, and MixedCurvPool.

The thresholds aligned with the center of the histogram seem to achieve the highest scores. This implies that aggregating too many nodes and too few nodes both have detrimental effects on the achieved results. Selecting the correct threshold is a balancing act. For new datasets, we would recommend starting with thresholds that split the dataset into two halves since those seem to present a very good baseline for initial experiments.

In terms of the different CurvPool variations, there is no clear winner. While HighCurvPool tends to achieve the highest scores across the board, it seems to be more sensitive to the choices of the threshold. MixedCurvPool is more consistent with fewer outliers in both directions. MixedCurvPool probably does best on datasets where the curvature values are distributed over a larger area. This explains its weak performance on the HIV dataset. Generally speaking, all three variations seem to be competitive, with differing strengths and weaknesses corresponding to individual datasets and thresholds.

A comparison of the different aggregation schemes is presented in Table~\ref{table:Ergebnisse_agg}. For CurvPool, summing all node representations in a pool is clearly on top for all the different datasets. Meanwhile, CliquePool tends to achieve the best results when averaging. Especially notable are the differences between summing and averaging on the Artificial dataset. 
We explain the success of the sum aggregation by its similarity to more expressive message-passing schemes with respect to the Weisefeiler-Leman test. As pools of different sizes typically occur, the sum provides information about the number of nodes utilized in each pool. In contrast, the mean is unable to determine the number of nodes utilized for pooling, similar to its inability to determine the number of neighbors in message-passing operations~\cite{xu2018powerful}. Thus, the sum aggregation provides important additional structural features and offers increased expressivity.

\begin{table}[tb]
\setlength{\tabcolsep}{6pt}
\centering
\caption{A comparison of different aggregation schemes for CliquePool and the best performing~\ref{table:Ergebnisse_overall}.}
\begin{tabular}{l ccc ccc}
\toprule
\multirow{2}{*}{Accuracy} & \multicolumn{3}{c}{CliquePool} & \multicolumn{3}{c}{CurvPool} \\
\cmidrule(lr){2-4}\cmidrule(lr){5-7}
& Sum & Avg & Max & Sum & Avg & Max\\
\midrule
HIV&75.76&\textbf{78.33}&73.05&\textbf{80.06}&79.44&78.26\\
Proteins&74.36&\textbf{74.54}&72.72&\textbf{77.81}&75.45&76.54\\
IMDB-BINARY&67.40&\textbf{68.39}&63.59&\textbf{70.80}&69.20&60.20\\
Artificial&72.80&73.40&\textbf{73.60}&\textbf{74.10}&69.70&63.70\\
\bottomrule
\end{tabular}
\label{table:Ergebnisse_agg}
\end{table}

\subsection{Runtime}
\label{Run-Time}
The runtimes presented in Table~\ref{table:Ergebnisse_Laufzeit} largely align with the established considerations of Section~\ref{Complexity}. CliquePool and CurvPool can utilize the precalculation of the poolings to drastically reduce the runtime per epoch and outperform DiffPool. Between CliquePool and CurvPool, runtimes are very close and in some cases even faster than the basic GCN. This can be explained through the larger number of aggregated nodes and the resulting smaller adjacency matrices. Especially meaningful is the amount of edges since it impacts the calculation of the Balanced Forman curvature negatively. Though this effect should be largely limited to the precalculation and not affect the time per epoch.

\begin{table}[tb]
\setlength{\tabcolsep}{6pt}
\begin{center}
\centering
\caption{Runtime per method and dataset. Pre represents the duration for the precomputation of the poolings. Epoch is the training time per epoch. All times are in seconds.}
\begin{tabular}{c cc cc cc cc}
  \toprule
  \multirow{2}{*}{Runtime (s)} & \multicolumn{2}{c}{HIV} & \multicolumn{2}{c}{Proteins} & \multicolumn{2}{c}{IMDB-BINARY} & \multicolumn{2}{c}{Artificial} \\
  \cmidrule(lr){2-3}\cmidrule(lr){4-5}\cmidrule(lr){6-7}\cmidrule(lr){8-9}
  & Epoch & Pre & 
    Epoch & Pre & Epoch & Pre & Epoch & Pre \\
  \midrule
GCN & $\mathbf{1.6}$ & - & $\mathbf{2.0}$ & - & 0.7 & - & $\mathbf{1.0}$ & -\\
DiffPool & 9.5 & - & 47.7 & - & 2.0 & - & 4.0 & -\\
CliquePool & 2.4 & $\mathbf{19}$ & 4.0 & $\mathbf{15}$ & $\mathbf{0.5}$ & $\mathbf{3}$ & $\mathbf{1.0}$ & 114 \\
CurvPool & 3.0 & 29 & 5.0 & 28 & $\mathbf{0.5}$ & 9 & $\mathbf{1.0}$ & $\mathbf{100}$ \\
  \bottomrule
\end{tabular}
\label{table:Ergebnisse_Laufzeit}
\end{center}
\end{table}

\section{Conclusion and Future Work}
We introduced CurvPool, a novel pooling approach for graph neural networks that are designed to be effective against over-smoothing and over-squashing. It is based on the Balanced Forman curvature, which represents the connectivity between nodes. A high curvature value occurs for densely connected areas of the graph, which are prone to over-smoothing. Bottlenecks are located at edges with a low curvature value, leading to over-squashing. Our proposed methods HighCurvPool and LowCurvPool directly reduce these critical areas by combining exactly these nodes, resulting in a coarser graph and more effective message-passing. Simultaneously reducing both of these areas is done using our proposed MixedCurvPool.
The first outstanding quality of CurvPool is its flexibility. The curvature can be calculated for any graph and always leads to a meaningful pooling metric while still being an inherent attribute of the graph itself. Other approaches like CliquePool on the other hand are servery limited through the need for the existence of specific structures within the graph. A graph without cliques of appropriate size will not lead to a suitable pooling. Meanwhile pooling via DiffPool is almost completely independent of the graph structure itself since it uses an external pooling metric in the form of a clustering. This approach also requires additional knowledge about the graph or extensive hyperparameter tuning to select the fitting clique sizes.

Another important factor is the low theoretical and practical runtime complexity of CurvPool which is linear in the number of edges. As the curvature and thus the pooling can be precomputed, the additional time during training is almost negligible.

Our empirical results on several graph classification tasks show the effectiveness of our approach. CurvPool comes out slightly ahead while comparing classification accuracy for a bunch of different datasets. The experiments have also shown the viability of the different CurvPool variations in highlighting their strengths and weaknesses on specific datasets and parameter combinations.
We found HighCurvPool using the sum aggregation to accomplish the strongest results consistently. We explain this as sets of nodes connected by edges with high curvature are prone to over-smoothing. These carry redundant features which HighCurvPool then reduces to a single node. The sum aggregation allows our method to utilize structural properties of the graph as the resulting state is influenced by the number of nodes in each pool.   
In summary, this flexibility, its slightly improved classification accuracy, and its low runtime complexity make CurvPool a valuable alternative to the established pooling methods.


\paragraph{Limitations}
Limitations of CurvPool directly stem from limitations in the Balanced Forman curvature itself. While this strategy is very flexible, it may not perfectly align with the nodes that should be pooled in the optimal scenario. However, in case other strategies for ranking pairs of nodes emerge, these can be directly integrated into our method. CurvPool also does not consider node features, which might further enhance its effectiveness, albeit reducing its ability to precompute clusters.
Additionally, while our empirical results already cover diverse datasets, CurvPool can be evaluated for additional tasks and against other methods to ensure its generalizability. 

\paragraph{Future work}
Our work opens up several directions for future work. While this paper focused on graph classification, it could be extended to node classification tasks in order to combine distant information. During our work, we noticed that the current theory on over-smoothing and over-squashing is unfit for pooled graphs. Metrics like the Dirichlet energy are not designed for pooling operations, making it challenging to quantify whether a pooling step can reduce over-smoothing. Thus, novel metrics and theoretical investigations are needed. Similarly, the effect of pooling methods in general on the curvature needs to be better understood from a theoretical perspective.

\paragraph{Acknowledgements} This research has been funded by the Federal Ministry of Education and Research of Germany and the state of North-Rhine Westphalia as part of the Lamarr-Institute for Machine Learning and Artificial Intelligence and by the Federal Ministry of Education and Research of Germany under grant no. 01IS22094E WEST-AI.

\bibliographystyle{unsrtnat}
\bibliography{references}  

\begin{thebibliography}{48}
\providecommand{\natexlab}[1]{#1}
\providecommand{\url}[1]{\texttt{#1}}
\expandafter\ifx\csname urlstyle\endcsname\relax
  \providecommand{\doi}[1]{doi: #1}\else
  \providecommand{\doi}{doi: \begingroup \urlstyle{rm}\Url}\fi

\bibitem[Kipf and Welling(2016)]{kipf16}
Thomas~N Kipf and Max Welling.
\newblock Semi-supervised classification with graph convolutional networks.
\newblock \emph{arXiv preprint arXiv:1609.02907}, 2016.

\bibitem[Roth and Liebig(2022{\natexlab{a}})]{roth2022forecasting}
Andreas Roth and Thomas Liebig.
\newblock Forecasting unobserved node states with spatio-temporal graph neural
  networks.
\newblock In \emph{2022 IEEE International Conference on Data Mining Workshops
  (ICDMW)}, pages 740--747. IEEE, 2022{\natexlab{a}}.

\bibitem[Zhang et~al.(2018{\natexlab{a}})Zhang, Cui, Neumann, and
  Chen]{zhang2018end}
Muhan Zhang, Zhicheng Cui, Marion Neumann, and Yixin Chen.
\newblock An end-to-end deep learning architecture for graph classification.
\newblock In \emph{Proceedings of the AAAI conference on artificial
  intelligence}, volume~32, 2018{\natexlab{a}}.

\bibitem[Pan et~al.(2018)Pan, Hu, Long, Jiang, Yao, and
  Zhang]{pan2018adversarially}
Shirui Pan, Ruiqi Hu, Guodong Long, Jing Jiang, Lina Yao, and Chengqi Zhang.
\newblock Adversarially regularized graph autoencoder for graph embedding.
\newblock \emph{arXiv preprint arXiv:1802.04407}, 2018.

\bibitem[Nt and Maehara(2019)]{nt2019revisiting}
Hoang Nt and Takanori Maehara.
\newblock Revisiting graph neural networks: All we have is low-pass filters.
\newblock \emph{arXiv preprint arXiv:1905.09550}, 2019.

\bibitem[Oono and Suzuki(2019)]{oono2019graph}
Kenta Oono and Taiji Suzuki.
\newblock Graph neural networks exponentially lose expressive power for node
  classification.
\newblock \emph{arXiv preprint arXiv:1905.10947}, 2019.

\bibitem[Chen et~al.(2020)Chen, Lin, Li, Li, Zhou, and Sun]{chen20}
Deli Chen, Yankai Lin, Wei Li, Peng Li, Jie Zhou, and Xu~Sun.
\newblock Measuring and relieving the over-smoothing problem for graph neural
  networks from the topological view.
\newblock In \emph{Proceedings of the AAAI Conference on Artificial
  Intelligence}, volume~34, pages 3438--3445, 2020.

\bibitem[Alon and Yahav(2020)]{alon20}
Uri Alon and Eran Yahav.
\newblock On the bottleneck of graph neural networks and its practical
  implications.
\newblock \emph{arXiv preprint arXiv:2006.05205}, 2020.

\bibitem[Topping et~al.(2021)Topping, Di~Giovanni, Chamberlain, Dong, and
  Bronstein]{topping21}
Jake Topping, Francesco Di~Giovanni, Benjamin~Paul Chamberlain, Xiaowen Dong,
  and Michael~M Bronstein.
\newblock Understanding over-squashing and bottlenecks on graphs via curvature.
\newblock \emph{arXiv preprint arXiv:2111.14522}, 2021.

\bibitem[Li et~al.(2018)Li, Han, and Wu]{li2018deeper}
Qimai Li, Zhichao Han, and Xiao-Ming Wu.
\newblock Deeper insights into graph convolutional networks for semi-supervised
  learning.
\newblock In \emph{Proceedings of the AAAI conference on artificial
  intelligence}, volume~32, 2018.

\bibitem[Cai and Wang(2020)]{cai2020note}
Chen Cai and Yusu Wang.
\newblock A note on over-smoothing for graph neural networks.
\newblock \emph{arXiv preprint arXiv:2006.13318}, 2020.

\bibitem[Yan et~al.(2022)Yan, Hashemi, Swersky, Yang, and Koutra]{yan2022two}
Yujun Yan, Milad Hashemi, Kevin Swersky, Yaoqing Yang, and Danai Koutra.
\newblock Two sides of the same coin: Heterophily and oversmoothing in graph
  convolutional neural networks.
\newblock In \emph{2022 IEEE International Conference on Data Mining (ICDM)},
  pages 1287--1292. IEEE, 2022.

\bibitem[Roth and Liebig(2022{\natexlab{b}})]{roth2022transforming}
Andreas Roth and Thomas Liebig.
\newblock Transforming pagerank into an infinite-depth graph neural network.
\newblock In \emph{Joint European Conference on Machine Learning and Knowledge
  Discovery in Databases}, pages 469--484. Springer, 2022{\natexlab{b}}.

\bibitem[Ying et~al.(2018)Ying, You, Morris, Ren, Hamilton, and
  Leskovec]{ying18}
Zhitao Ying, Jiaxuan You, Christopher Morris, Xiang Ren, Will Hamilton, and
  Jure Leskovec.
\newblock Hierarchical graph representation learning with differentiable
  pooling.
\newblock \emph{Advances in neural information processing systems}, 31, 2018.

\bibitem[Luzhnica et~al.(2019)Luzhnica, Day, and Lio]{luzhnica19}
Enxhell Luzhnica, Ben Day, and Pietro Lio.
\newblock Clique pooling for graph classification.
\newblock \emph{arXiv preprint arXiv:1904.00374}, 2019.

\bibitem[Bianchi and Lachi(2023)]{bianchi2023expressive}
Filippo~Maria Bianchi and Veronica Lachi.
\newblock The expressive power of pooling in graph neural networks.
\newblock \emph{arXiv preprint arXiv:2304.01575}, 2023.

\bibitem[Forman(2003)]{forman03}
Forman.
\newblock Bochner's method for cell complexes and combinatorial ricci
  curvature.
\newblock \emph{Discrete \& Computational Geometry}, 29:\penalty0 323--374,
  2003.

\bibitem[Ollivier(2007)]{ollivier07}
Yann Ollivier.
\newblock Ricci curvature of metric spaces.
\newblock \emph{Comptes Rendus Mathematique}, 345\penalty0 (11):\penalty0
  643--646, 2007.

\bibitem[Ollivier(2009)]{ollivier09}
Yann Ollivier.
\newblock Ricci curvature of markov chains on metric spaces.
\newblock \emph{Journal of Functional Analysis}, 256\penalty0 (3):\penalty0
  810--864, 2009.

\bibitem[Yuan and Ji(2020)]{yuan20}
Hao Yuan and Shuiwang Ji.
\newblock Structpool: Structured graph pooling via conditional random fields.
\newblock In \emph{Proceedings of the 8th International Conference on Learning
  Representations}, 2020.

\bibitem[Zhang et~al.(2018{\natexlab{b}})Zhang, Cui, Neumann, and
  Chen]{zhang18}
Muhan Zhang, Zhicheng Cui, Marion Neumann, and Yixin Chen.
\newblock An end-to-end deep learning architecture for graph classification.
\newblock In \emph{Proceedings of the AAAI conference on artificial
  intelligence}, volume~32, 2018{\natexlab{b}}.

\bibitem[Ranjan et~al.(2020)Ranjan, Sanyal, and Talukdar]{ranjan20}
Ekagra Ranjan, Soumya Sanyal, and Partha Talukdar.
\newblock Asap: Adaptive structure aware pooling for learning hierarchical
  graph representations.
\newblock In \emph{Proceedings of the AAAI Conference on Artificial
  Intelligence}, volume~34, pages 5470--5477, 2020.

\bibitem[Lee et~al.(2019)Lee, Lee, and Kang]{lee19}
Junhyun Lee, Inyeop Lee, and Jaewoo Kang.
\newblock Self-attention graph pooling.
\newblock In \emph{International conference on machine learning}, pages
  3734--3743. PMLR, 2019.

\bibitem[Wang et~al.(2020)Wang, Li, Ma, Montufar, Zhuang, and Fan]{wang20}
Yu~Guang Wang, Ming Li, Zheng Ma, Guido Montufar, Xiaosheng Zhuang, and Yanan
  Fan.
\newblock Haar graph pooling.
\newblock In \emph{International conference on machine learning}, pages
  9952--9962. PMLR, 2020.

\bibitem[Gao et~al.(2019)Gao, Chen, and Ji]{gao19}
Hongyang Gao, Yongjun Chen, and Shuiwang Ji.
\newblock Learning graph pooling and hybrid convolutional operations for text
  representations.
\newblock In \emph{The world wide web conference}, pages 2743--2749, 2019.

\bibitem[Diehl et~al.(2019)Diehl, Brunner, Le, and Knoll]{diehl19}
Frederik Diehl, Thomas Brunner, Michael~Truong Le, and Alois Knoll.
\newblock Towards graph pooling by edge contraction.
\newblock In \emph{ICML 2019 workshop on learning and reasoning with
  graph-structured data}, 2019.

\bibitem[Li et~al.(2020)Li, Ma, Wang, Aggarwal, Wang, and Tang]{li20}
Juanhui Li, Yao Ma, Yiqi Wang, Charu Aggarwal, Chang-Dong Wang, and Jiliang
  Tang.
\newblock Graph pooling with representativeness.
\newblock In \emph{2020 IEEE International Conference on Data Mining (ICDM)},
  pages 302--311. IEEE, 2020.

\bibitem[Nguyen and Grishman(2018)]{nguyen18}
Thien Nguyen and Ralph Grishman.
\newblock Graph convolutional networks with argument-aware pooling for event
  detection.
\newblock In \emph{Proceedings of the AAAI Conference on Artificial
  Intelligence}, volume~32, 2018.

\bibitem[Zhang et~al.(2021)Zhang, Satapathy, Guttery, G{\'o}rriz, and
  Wang]{zhang21}
Yu-Dong Zhang, Suresh~Chandra Satapathy, David~S Guttery, Juan~Manuel
  G{\'o}rriz, and Shui-Hua Wang.
\newblock Improved breast cancer classification through combining graph
  convolutional network and convolutional neural network.
\newblock \emph{Information Processing \& Management}, 58\penalty0
  (2):\penalty0 102439, 2021.

\bibitem[Du et~al.(2021)Du, Wang, Miao, and Zhang]{du21}
Jinlong Du, Senzhang Wang, Hao Miao, and Jiaqiang Zhang.
\newblock Multi-channel pooling graph neural networks.
\newblock In \emph{IJCAI}, pages 1442--1448, 2021.

\bibitem[Lei et~al.(2022)Lei, Liu, Jiang, Liao, Cai, and Zhao]{lei22}
Fangyuan Lei, Xun Liu, Jianjian Jiang, Liping Liao, Jun Cai, and Huimin Zhao.
\newblock Graph convolutional networks with higher-order pooling for
  semisupervised node classification.
\newblock \emph{Concurrency and Computation: Practice and Experience},
  34\penalty0 (16):\penalty0 e5695, 2022.

\bibitem[Zhao et~al.(2022)Zhao, Xie, and Wang]{zhao22}
Hongyu Zhao, Jiazhi Xie, and Hongbin Wang.
\newblock Graph convolutional network based on multi-head pooling for short
  text classification.
\newblock \emph{IEEE Access}, 10:\penalty0 11947--11956, 2022.

\bibitem[Noutahi et~al.(2019)Noutahi, Beaini, Horwood, Gigu{\`e}re, and
  Tossou]{noutahi2019towards}
Emmanuel Noutahi, Dominique Beaini, Julien Horwood, S{\'e}bastien Gigu{\`e}re,
  and Prudencio Tossou.
\newblock Towards interpretable sparse graph representation learning with
  laplacian pooling.
\newblock \emph{arXiv preprint arXiv:1905.11577}, 2019.

\bibitem[Bianchi et~al.(2020)Bianchi, Grattarola, and
  Alippi]{bianchi2020spectral}
Filippo~Maria Bianchi, Daniele Grattarola, and Cesare Alippi.
\newblock Spectral clustering with graph neural networks for graph pooling.
\newblock In \emph{International conference on machine learning}, pages
  874--883. PMLR, 2020.

\bibitem[Khasahmadi et~al.(2020)Khasahmadi, Hassani, Moradi, Lee, and
  Morris]{khasahmadi2020memory}
Amir~Hosein Khasahmadi, Kaveh Hassani, Parsa Moradi, Leo Lee, and Quaid Morris.
\newblock Memory-based graph networks.
\newblock \emph{arXiv preprint arXiv:2002.09518}, 2020.

\bibitem[Liu et~al.(2021)Liu, Jian, Li, Zhang, Lai, and
  Xu]{liu2021hierarchical}
Ning Liu, Songlei Jian, Dongsheng Li, Yiming Zhang, Zhiquan Lai, and Hongzuo
  Xu.
\newblock Hierarchical adaptive pooling by capturing high-order dependency for
  graph representation learning.
\newblock \emph{IEEE Transactions on Knowledge and Data Engineering}, 2021.

\bibitem[Morris et~al.(2019)Morris, Ritzert, Fey, Hamilton, Lenssen, Rattan,
  and Grohe]{Morris_Ritzert_Fey_Hamilton_Lenssen_Rattan_Grohe_2019}
Christopher Morris, Martin Ritzert, Matthias Fey, William~L. Hamilton, Jan~Eric
  Lenssen, Gaurav Rattan, and Martin Grohe.
\newblock Weisfeiler and leman go neural: Higher-order graph neural networks.
\newblock \emph{Proceedings of the AAAI Conference on Artificial Intelligence},
  33:\penalty0 4602--4609, 2019.
\newblock \doi{10.1609/aaai.v33i01.33014602}.

\bibitem[Fey et~al.(2020)Fey, Yuen, and Weichert]{fey2020hierarchical}
Matthias Fey, Jan-Gin Yuen, and Frank Weichert.
\newblock Hierarchical inter-message passing for learning on molecular graphs.
\newblock \emph{arXiv preprint arXiv:2006.12179}, 2020.

\bibitem[Bron and Kerbosch(1973)]{bron73}
Coen Bron and Joep Kerbosch.
\newblock Algorithm 457: finding all cliques of an undirected graph.
\newblock \emph{Communications of the ACM}, 16\penalty0 (9):\penalty0 575--577,
  1973.

\bibitem[Tomita et~al.(2006)Tomita, Tanaka, and Takahashi]{tomita06}
Etsuji Tomita, Akira Tanaka, and Haruhisa Takahashi.
\newblock The worst-case time complexity for generating all maximal cliques and
  computational experiments.
\newblock \emph{Theoretical computer science}, 363\penalty0 (1):\penalty0
  28--42, 2006.

\bibitem[Eppstein et~al.(2013)Eppstein, L{\"o}ffler, and Strash]{eppstein13}
David Eppstein, Maarten L{\"o}ffler, and Darren Strash.
\newblock Listing all maximal cliques in large sparse real-world graphs.
\newblock \emph{Journal of Experimental Algorithmics (JEA)}, 18:\penalty0 3--1,
  2013.

\bibitem[Blakely et~al.(2021)Blakely, Lanchantin, and Qi]{blakely21}
Derrick Blakely, Jack Lanchantin, and Yanjun Qi.
\newblock Time and space complexity of graph convolutional networks.
\newblock \emph{Accessed on: Dec}, 31, 2021.

\bibitem[Wu et~al.(2018)Wu, Ramsundar, Feinberg, Gomes, Geniesse, Pappu,
  Leswing, and Pande]{wu18}
Zhenqin Wu, Bharath Ramsundar, Evan~N Feinberg, Joseph Gomes, Caleb Geniesse,
  Aneesh~S Pappu, Karl Leswing, and Vijay Pande.
\newblock Moleculenet: a benchmark for molecular machine learning.
\newblock \emph{Chemical science}, 9\penalty0 (2):\penalty0 513--530, 2018.

\bibitem[Borgwardt et~al.(2005)Borgwardt, Ong, Schönauer, Vishwanathan, Smola,
  and Kriegel]{10.1093/bioinformatics/bti1007}
Karsten~M. Borgwardt, Cheng~Soon Ong, Stefan Schönauer, S.~V.~N. Vishwanathan,
  Alex~J. Smola, and Hans-Peter Kriegel.
\newblock {Protein function prediction via graph kernels}.
\newblock \emph{Bioinformatics}, 21:\penalty0 i47--i56, 06 2005.
\newblock \doi{10.1093/bioinformatics/bti1007}.

\bibitem[Yanardag and Vishwanathan(2015)]{yanardag2015deep}
Pinar Yanardag and SVN Vishwanathan.
\newblock Deep graph kernels.
\newblock In \emph{Proceedings of the 21th ACM SIGKDD international conference
  on knowledge discovery and data mining}, pages 1365--1374, 2015.

\bibitem[Watts(1999)]{watts99}
Duncan~J Watts.
\newblock Networks, dynamics, and the small-world phenomenon.
\newblock \emph{American Journal of sociology}, 105\penalty0 (2):\penalty0
  493--527, 1999.

\bibitem[Ioffe and Szegedy(2015)]{ioffe15}
Sergey Ioffe and Christian Szegedy.
\newblock Batch normalization: Accelerating deep network training by reducing
  internal covariate shift.
\newblock In \emph{International conference on machine learning}, pages
  448--456. pmlr, 2015.

\bibitem[Xu et~al.(2018)Xu, Hu, Leskovec, and Jegelka]{xu2018powerful}
Keyulu Xu, Weihua Hu, Jure Leskovec, and Stefanie Jegelka.
\newblock How powerful are graph neural networks?
\newblock \emph{arXiv preprint arXiv:1810.00826}, 2018.

\end{thebibliography}






\end{document}